%% file: fan.tex
\documentclass{article}
\usepackage{ijcai17}
\input{mystyle}

\title{Robust Localized Multi-view Subspace Clustering}
\author{Yanbo Fan$^{1,3}$
	, Jian Liang$^{1,2,3}$ 
	, Ran He$^{1,2,3}$ 
	, Bao-Gang Hu$^{1,3}$ 
	, Siwei Lyu$^{4}$\\
	{$^{1}$National Laboratory of Pattern Recognition, CASIA}\\
	{$^{2}$Center for Research on Intelligent Perception and Computing, CASIA}\\
	{$^{3}$University of Chinese Academy of Sciences (UCAS)}\\
	{$^{4}$Department of Computer Science, University at Albany, SUNY} \\
	{\{yanbo.fan, jian.liang, rhe, hubg\}@nlpr.ia.ac.cn}, slyu@albany.edu
}

\begin{document}

\maketitle

\input{paper/abstract}

\input{paper/introduction}

\input{paper/related_work}

\input{paper/proposed_method}

\input{paper/experiments}

\input{paper/conclusion}

\input{paper/appendix}

\bibliographystyle{named}
\bibliography{ijcai17}

\end{document}

%% file: mystyle.tex
\usepackage{times}
\usepackage{helvet}
\usepackage{courier}
\usepackage{graphicx}
\usepackage{subfigure}
\usepackage{amssymb}
\usepackage{amsmath}
\usepackage{array}
\usepackage{multirow}
\usepackage{epsfig}
\usepackage{epstopdf}
\usepackage{color}

\usepackage{algorithm} 
\usepackage{algorithmic} 

\def\bZ{\mathbf{Z}}
\def\bS{\mathbf{S}}
\def\bY{\mathbf{Y}}

\def\bX{\mathbf{X}}
\def\bP{\mathbf{P}}
\def\bI{\mathbf{I}}
\def\b1{\mathbf{1}}
\def\by{\mathbf{y}}
\def\bt{\mathbf{t}}
\def\bp{\mathbf{p}}

\usepackage{footmisc}

%% file: paper/abstract.tex
\begin{abstract}
In multi-view clustering, different views may have different confidence levels when learning a consensus representation. 
Existing methods usually address this by assigning distinctive weights to different views. 
However, due to noisy nature of real-world applications, the confidence levels of samples in the same view may also vary. Thus considering a unified weight for a view may lead to suboptimal solutions. 
In this paper, we propose a novel localized multi-view subspace clustering model that considers the confidence levels of both views and samples. 
By assigning weight to each sample under each view properly, we can obtain a robust consensus representation via fusing the noiseless structures among views and samples.
We further develop a regularizer on weight parameters based on the convex conjugacy theory, and samples weights are determined in an adaptive manner.
An efficient iterative algorithm is developed with a convergence guarantee. Experimental results on four benchmarks demonstrate the correctness and effectiveness of the proposed model.
\end{abstract}

%% file: paper/introduction.tex
\section{Introduction}
Many real-world data have representations in multiple feature types and modalities. For example, images can be represented by different types of features such as SIFT or HOG, and web pages usually consist of hyperlinks and texts. Each such representation is usually referred to as a view of a multi-view dataset. As different views are often collected from diverse domains using different feature extractors, they may contain information complementary to other views.  As such, multi-view data can provide richer and more relevant information comparing to single-view data. Accordingly, multi-view data processing has become a popular topic in machine learning and pattern recognition \cite{xu2016discriminatively,sun2015multi,cao2015diversity,zhao2015dual}.

Multi-view clustering is an unsupervised multi-view data analysis problem, which aims to group a multi-view dataset into distinct clusters.  
One common strategy for multi-view clustering is to learn a consensus representation that can reflect the latent clustering structure shared by different views \cite{liu2016multiple,gao2015multi,xu2015multi,xu2013survey,liu2013multi}.
As multi-view data are often collected from diverse domains and the statistical properties among views may vary, different views may have different confidence levels when pursuing a consensus representation (here the confidence level of a view or a sample refers to its ability to uncover the underlying clustering structures). For example, a view that potentially has good clustering structures should have higher confidence level and vice-versa. 
Existing works usually address this by assigning distinctive weights to different views (e.g., assign larger weights to views that have higher confidence levels), either manually or automatically \cite{xu2016discriminatively,wang2014multi,wang2013grassmannian,xia2010multiview}. 
%
However, due to the complexity and noisy nature of multi-view data in real world applications, the confidence levels of samples in the same view may also vary (e.g., outliers or noisy samples should have lower confidence level comparing to the uncorrupted ones in the same view). 
%
Thus considering a unified weight for a view may result in suboptimal solutions.  

In this paper, we propose a novel robust localized multi-view subspace clustering model that considers the confidence levels of both samples and views when learning a consensus representation. Specifically, we introduce a nonnegative weight parameter to each sample under each view. By assigning weights to samples under each view properly (e.g., assign smaller weights to samples with lower confidence levels), we can obtain a robust consensus representation via fusing the noiseless structures among different views and samples. 
Meanwhile, as we usually do not have prior information about samples and views confidence levels in unsupervised situation, it may be impractical to manually assign weight to each sample, especially for large dataset. We thus design a regularizer on weight parameters based on the convex conjugacy theory, and sample weights are adapted during the optimization. We then develop an efficient iterative algorithm to solve this problem, and perform extensive experiments on four benchmarks to demonstrate the correctness and effectiveness of the proposed method.

The main contributions of our work are: 
(1) We propose to consider the confidence levels of both views and samples when learning a consensus representation for multi-view clustering. 
(2) We design a regularizer on samples weights so that they are adaptively assigned during the optimization process. The learned weights can reflect samples confidence levels to some extent. 
The whole formulation is easy to optimize and the convergence is guaranteed.
(3) The proposed model outperforms the related state-of-the-art multi-view clustering methods on four real-world multi-view datasets.

%% file: paper/related_work.tex
\section{Related Work}

Recently, many spectral-based subspace clustering methods have been proposed \cite{hu2014smooth,feng2014robust,he2014robust,elhamifar2013sparse,liu2013robust,zhang2013robust}. 
These approaches first learn a similarity matrix $\bS$ based on data's self-representation. They then obtain clustering results by applying spectral clustering algorithm like \cite{shi2000normalized} on the learned similarity matrix.
For example, Sparse Subspace Clustering (SSC) \cite{elhamifar2013sparse} constructs the similarity matrix $\bS$ through finding a sparse linear representation of each sample. Low-Rank Representation (LRR) \cite{liu2013robust} emphasizes low-rank property on data's self-representation during the optimization. 
Though proved to be effective in many applications, these methods mainly focus on single-view data. 

To explore the complementary information involved in multi-view data, several multi-view clustering algorithms have been proposed \cite{lu2016convex,li2016multiple,wang2013multi,kumar2011co,kumar2011co_training}.
For example, \cite{kumar2011co} propose a co-regularization framework to regularize the difference between views  Laplacian embedding. \cite{liu2013multi} propose a multi-view nonnegative matrix factorization framework to pursue a consensus representation.  
To extend single-view subspace clustering algorithm to multi-view situation, \cite{cao2015diversity} propose a diversity-induced multi-view subspace model that explicitly enforces the diversity of different representations. \cite{gao2015multi} propose to learn a consistence clustering structure and the subspace representation of each view simultaneously. Instead of considering each view equally, \cite{xia2010multiview} propose to learn an optimal weighting to linearly combine different views representations. \cite{wang2014multi} develop a minimax optimization framework to minimize the loss of the worst case with maximum disagreements.  
However, these methods usually learn a unified weight for a view and do not consider the confidence level variations of samples in the same view. 
Recently, \cite{gonen2014localized} propose a localized multiple kernel learning framework that considers sample-specific weights to different kernels. However, the kernels are pre-specified in their work and they further constrain that the weights should sum to one for each sample. 
%
In contrast, we propose to learn samples weights and consensus representation jointly and develop a novel weighting strategy based on convex conjugacy theory.

%% file: paper/proposed_method.tex
\section{The Proposed Method}


\input{paper/preliminary}

\input{paper/proposed_formulation}

\input{paper/discuss_psi}
\input{paper/optimization}

%% file: paper/preliminary.tex
\subsection{Terms and Notations}
We use $||\cdot||_F$, $||\cdot||_1$, $||\cdot||_2$, $||\cdot||_{2,1}$ to denote \emph{Frobenius}-norm, the $\ell_1$-norm (sum of absolute value), the $\ell_2$-norm and the $\ell_{2,1}$-norm (sum of of the $\ell_2$ norm of columns of a matrix). $|\cdot|$ takes the absolute value of the elements in a matrix. 
We denote $R_+ = \{t| t \geq 0\}$. 
For a vector $\by \in R^{n}$, $diag(\by)$ denotes a square diagonal matrix with elements of $\by$ on its main diagonal, and $\by_i$ is its $i$-th entry. 
For a matrix $\bY \in R^{n \times n}$, $diag(\bY)$ is the vector formed from its main diagonal elements.
We use $\bY_{\cdot i}$ and $\bY_{j\cdot}$ to denote the $i$-th column and $j$-th row of matrix $\bY$, respectively. 
$\bY_{ij}$ denotes the entry of $\bY$ at $i$-th row and $j$-th column.

%% file: paper/proposed_formulation.tex
\subsection{Proposed Formulation}
In this section, we will introduce a novel approach to extend the single-view subspace clustering model to multi-view situation. Specifically, we will demonstrate our main ideas based on SSC model due to its good interpretability and effectiveness \cite{elhamifar2013sparse}. 
Given a multi-view dataset
\(\bX = [\bX^{1};\bX^{2};\dots;\bX^{m}] \in R^{d \times n}\) with $m$ views sampled from $c$ clusters, $n$ is the number of total samples. 
\(\bX^{v} \in R^{d_{v} \times n}\) is the feature matrix corresponding to the $v$-th view and $d_{v}$ is its feature, $d = \sum_{v=1}^{m} d_{v}$. 
Single-view SSC model corresponding to $v$-th view can be described as
\begin{equation}
\min_{\bZ^v \in R^{n \times n}} ||\bX^v - \bX^v \bZ^v||_F^2 + \beta ||\bZ^v||_1, \ diag(\bZ^v) = 0.
\label{naive_ssc}
\end{equation}
where $\bZ^v \in R^{n \times n}$ is the learned sparse self-representation of $v$-th view and $\bZ^v_{\cdot i}$ is the self-representation of $i$-th sample under $v$-th view, $\beta$ is a nonnegative parameter that trades off the reconstruction error and the sparse constraint $||\bZ^v||_1$. After getting the optimal $\bZ^v$, \(\bS = (|\bZ^v| + |(\bZ^{v})^T|)/2\) is used to build a similarity matrix, and the final clustering results of $v$-th view is obtained by applying spectral clustering on $\bS$.

To explore the complementary information contained in multi-view data, a direct approach is to perform single-view subspace clustering model (\ref{naive_ssc}) on each view separately and then build similarity matrix $\bS$ as $\bS = \frac{1}{m} \sum_{v=1}^{m} \frac{|\bZ^v| + |(\bZ^{v})^T|}{2}$.
However, this na\"ive extension cannot make use of the complementary information in multi-view data very well as 1) it considers each view independently and 2) it simply treats all views and samples in an equal manner.  
As multi-view data are often collected from diverse domains by different types of feature extractors and noise is inevitable in real-world applications, different views usually have different confidence levels and the confidence levels of samples in the same view may also vary (e.g., outliers or noisy samples should have low confidence levels). Thus na\"ive treating each view equally or considering a unified weight for a view may both lead to suboptimal solutions. To better explore the complementary information in multi-view data, we propose a robust localized multi-view subspace clustering model to consider the confidence levels of both views and samples as
\begin{equation}
\small
\begin{aligned}
& \min_{\{\bZ^v\}_{v=1}^m, \bZ^*,\bP} \ \sum_{v=1}^{m} \sum_{j=1}^{n} \{ \bP_{vj} (||\bX_{\cdot j}^v - \bX^v \bZ_{\cdot j}^v||_F^2 + \lambda ||\bZ_{\cdot j}^v - \bZ_{\cdot j}^*||_F^2)  \\ 
& \quad \qquad \qquad \qquad \qquad + \psi(\bP_{vj}) \} + \beta||\bZ^*||_1 \\
& \ \ s.t. \quad \bP_{vj} \geq 0, \ diag(\bZ^v) = 0, v=1,\dots,m, \ j = 1, \dots, n. \\
\end{aligned}
\label{proposed_model}
\end{equation}
where $\bZ^* \in R^{n \times n}$ is the learned sparse consensus representation, and $(\lambda,\beta)$ are two nonnegative trade-off parameters. 
In model (\ref{proposed_model}), we explicitly introduce a nonnegative weight parameter for each sample under each view to reflect its confidence level. 
$\bP \in R^{m \times n}$ denotes the weight matrix and $\bP_{vj}$ is the weight of $j$-th sample under $v$-th view. 
By properly assigning weight to each sample under each view (e.g., assign small weights to outliers and noisy samples), $\bZ^*$ in model (\ref{proposed_model}) can better explore the complementary information contained in each $\bZ^v$. 
As we usually do not have prior information about the confidence levels of samples and views in unsupervised setting, it is impractical to obtain sample weights $\bP$ manually. Thus we further incorporate the updating of $\bP$ into the model optimization with a regularizer $\psi(\cdot)$ on $\bP$, and $\bP$ is adapted during the model optimization.

%% file: paper/discuss_psi.tex
\input{paper/fig_latent_minimizer}

\subsection{Discussion of $\psi(\cdot)$}
\label{discuss_regularizer}
One key issue in model (\ref{proposed_model}) is to design a proper regularizer $\psi(\cdot)$, which determines the updating of weight parameters. 
In order to obtain a robust consensus representation $\bZ^*$ among views and samples, outliers or noisy samples under each view should be assigned with smaller weights and samples with good local structures should obtain relative larger weights. 
Denote the loss of $j$-th sample under $v$-th view as $\ell_{vj} = ||\bX_{\cdot j}^v - \bX^v \bZ_{\cdot j}^v||_F^2 + \lambda ||\bZ_{\cdot j}^v - \bZ_{\cdot j}^*||_F^2$.  Model (\ref{proposed_model}) with respect to $\bP_{vj}$ becomes
\begin{equation}
\min_{\bP_{vj} \geq 0} \ \bP_{vj} \ell_{vj} + \psi(\bP_{vj})
\label{optimal_p}
\end{equation}
Based on the convex conjugacy theory \cite{boyd2004convex}, we have 

\textbf{Lemma 1} Problem (\ref{optimal_p}) is related to a certain latent function $\phi(\ell) = -\psi^*(-\ell)$, 
where $\psi^*(\cdot)$ denotes the convex conjugate of function $\psi(\cdot)$ 
\footnote{
The convex conjugate of function $f(\bt)$ is defined as 
$f^*(\by) = \sup_{\bt \in dom f} \left( \by^T\bt - f(\bt) \right )$ \cite{boyd2004convex}. 
}.

\emph{Proof. Based on the convex conjugacy theory, we have 
$\phi(\ell) = -\psi^*(-\ell) = -\max_{v \geq 0} \{-v\ell - \psi(v)\} = \min_{v \geq 0} \{v\ell + \psi(v)\}$. 
$\phi(\ell)$ is concave due to the fact that the convex conjugate $\psi^*(\ell)$ is convex.
}

This on the other side encourages us to design regularizer $\psi(\cdot)$ based on the latent loss functions. Given a concave loss function $\phi(\ell): R_+ \rightarrow R_+$, we can define 
$\psi(p) = \max_{\ell \geq 0} \{-p\ell + \phi(\ell)\}$, and $\phi(\ell) = \min_{p \geq 0} \{p\ell + \psi(p)\}$ accordingly.
For instance, we can define $\psi(\cdot): R_{+} \rightarrow R_{+}$ as 
\begin{equation}
\psi(p) = \gamma p + \frac{1}{p} - 2
\label{l1_l2_regularizer}
\end{equation}
where $\gamma$ is a nonnegative hyper-parameter, $\psi(p)$ is a convex function for $p > 0$. Its corresponding latent loss function is $\phi(\ell) = 2(\sqrt{\gamma + \ell} - 1)$. 
By substituting eq (\ref{l1_l2_regularizer}) into problem (\ref{optimal_p}), the optimal weight for $j$-th sample under $v$-th view in (\ref{optimal_p}) is calculated as 
\begin{equation}
\bP_{vj}^* = \sigma(\ell_{vj}) = \frac{1}{\sqrt{\gamma + \ell_{vj}}}
\end{equation} 
where $\sigma(\ell)$ is named as minimizer function. 
Figure \ref{latent_minimizer} gives a graphic illustration of $\phi(\ell)$ and $\sigma(\ell)$. 
We can tell that $\sigma(\ell)$ is monotone decreasing with respect to loss $\ell$, and samples with larger losses will get smaller weights. As outliers or noisy samples are usually considered to be deviated from the major normal ones, they can usually cause larger reconstruction loss and will get smaller weights accordingly. Meanwhile, samples with good local structures probably can be reconstructed very well and will get larger weights consequently.
Through this, the learned weight matrix $\bP$ can reflect the confidence levels of samples under each view.
Its correctness and effectiveness are further demonstrated in the experimental part. 
By substituting eq (\ref{l1_l2_regularizer}) into model (\ref{proposed_model}), we obtain the proposed robust multi-view subspace clustering model.

%% file: paper/fig_latent_minimizer.tex
\begin{figure}[t]
	\centering
	\subfigure[Latent loss function $\phi(\ell)$]{
		\includegraphics[height=1.23in]{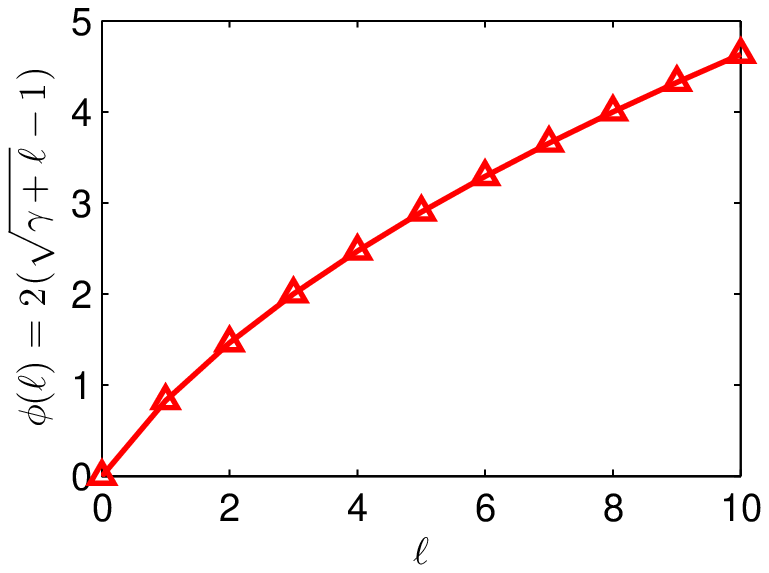}
	}
	\subfigure[Minimizer function $\sigma(\ell)$]{
		\includegraphics[height=1.23in]{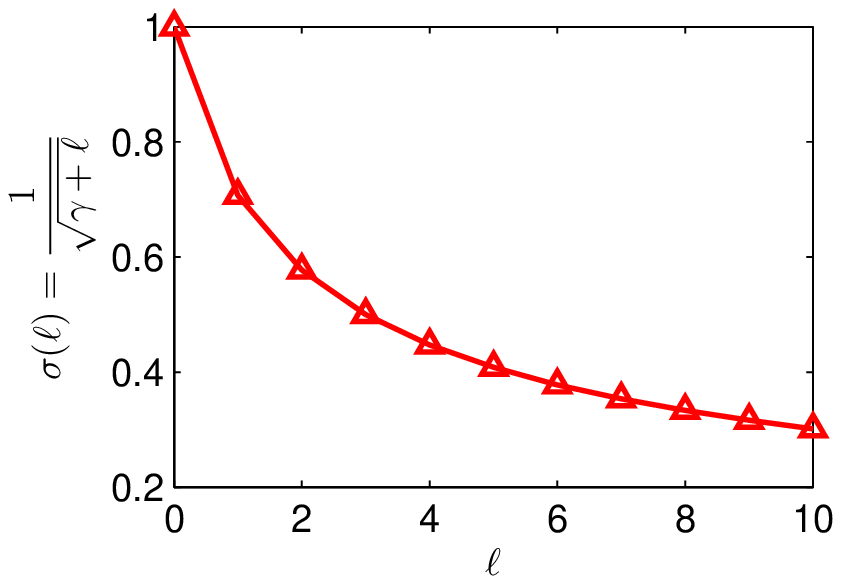}
	}
	\caption{Graphical representations of latent loss function and its corresponding minimizer function, $\gamma = 1$.}
	\label{latent_minimizer}
\end{figure}

%% file: paper/optimization.tex
\section{Optimization}

\subsection{Optimization Algorithm}
Although problem (\ref{proposed_model}) is not jointly convex for all variables, it is convex with respect to each variable while fixing the others. Thus we can develop an efficient block coordinate descent algorithm for it. The overall algorithm is summarized in Algorithm 1. Specifically, the variables $\bP$, $\{\bZ^v\}_{v=1}^m$ and $\bZ^*$ are updated as follows:
\\ \\
\textbf{$\bP$-step}
Update weight matrix $\bP$ with fixed $\{\{\bZ^v\}_{v=1}^m, \bZ^*\}$. 
According to the discussion in  Section \ref{discuss_regularizer}, the optimizing of each $\bP_{vj}$ refers to a convex optimization problem (\ref{optimal_p}) and it has a closed form solution as 
\begin{equation}
\bP_{vj} = \frac{1}{\sqrt{\gamma + ||\bX_{\cdot j}^v - \bX^v \bZ_{\cdot j}^v||_F^2 + \lambda ||\bZ_{\cdot j}^v - \bZ_{\cdot j}^*||_F^2}}, 
\label{update_p}
\end{equation}
where $1 \leq v \leq m$, $1 \leq j \leq n$.
\\ \\
\textbf{$\bZ^v$-step}
Update each $\bZ^v$ while fixing $\{\bZ^*,\bP\}$. The optimization of $\bZ^v$ is equivalent to 
\begin{equation}
\begin{aligned}
& \bZ^v = \arg \min_{\bZ^v} ||\bX^v - \bX^v \bZ^v||_F^2  + \lambda ||\bZ^v - \bZ^*||_F^2 \\
& s.t. \ \ diag(\bZ^v) = 0.
\end{aligned}
\end{equation} 
Its Lagrange function is 
\begin{equation}
\small
L(\bZ^v, \by) =  ||\bX^v - \bX^v \bZ^v||_F^2  + \lambda ||\bZ^v - \bZ^*||_F^2 + \langle \by, diag(\bZ^v) \rangle,
\end{equation}
where $\by \in R^{n}$ is Lagrange multiplier and
$\langle \cdot, \cdot \rangle$ denotes the inner product of two vectors. 
By setting the derivative of $L(\bZ^v, \by)$ with respect to $\bZ^v$ and $\by$ to zero, we have
\begin{equation}
\small
\left \{
\begin{split}
& (\bX^v)^T\bX^v\bZ^v - (\bX^v)^T\bX^v + \lambda(\bZ^v - \bZ^*) + diag(\frac{\by}{2}) = 0, \\
& diag(\bZ^v) = 0.
\end{split}
\right.
\end{equation}
Therefore, $\bZ^v$ can be obtained as
\begin{equation}
\bZ^v = ((\bX^v)^T\bX^v + \lambda \bI)^{-1}((\bX^v)^T\bX^v+\lambda \bZ^* - diag(\frac{\by}{2})),
\label{update_z_v}
\end{equation}
where $\bI$ is a $n \times n$ identity matrix and $\by_{i} = \frac{2\langle[(\bX^v)^T\bX^v + \lambda \bI)^{-1}]_{i\cdot} \ , \ [(\bX^v)^T\bX^v+\lambda \bZ^*]_{\cdot i}\rangle}{[(\bX^v)^T\bX^v + \lambda \bI)^{-1}]_{ii}}$.
\\ \\
\textbf{$\bZ^*$-step} 
Update $\bZ^*$ while fixing $\{\{\bZ^v\}_{v=1}^m, \bP\}$. Problem (\ref{proposed_model}) becomes 
\begin{equation}
\bZ^* = \arg\min_{\bZ^*} \lambda \sum_{v=1}^{m} \sum_{j=1}^{n}  \bP_{vj}||\bZ_{\cdot j}^* - \bZ_{\cdot j}^v||_F^2 + \beta||\bZ^*||_1
\label{problem_z_*}
\end{equation}
It obtains a unique solution as
\begin{equation}
\bZ_{ij}^* = \left \{
\begin{split}
&  \frac{2\lambda \sum_{v=1}^{m}\bP_{vj} \bZ^v_{ij} - \beta}{2\lambda \sum_{v=1}^{m}\bP_{vj}}, \quad if \ \sum_{v=1}^{m}\bP_{vj} \bZ^v_{ij} > \frac{\beta}{2\lambda} \\
&  \frac{2\lambda \sum_{v=1}^{m}\bP_{vj} \bZ^v_{ij} + \beta}{2\lambda \sum_{v=1}^{m}\bP_{vj}}, \quad if \ \sum_{v=1}^{m}\bP_{vj} \bZ^v_{ij} < \frac{-\beta}{2\lambda} \\
& \; 0. \quad \; \qquad \qquad \qquad \qquad otherwise
\end{split}
\right.
\label{update_z_*}
\end{equation}
where $1 \leq i \leq n$, $1 \leq j \leq n$. 

Seen from eq (\ref{problem_z_*}) and eq (\ref{update_z_*}), the optimal $\bZ^*$ is a weighted average over each view's self-representation with a sparse constraint. 
As we have discussed in Section \ref{discuss_regularizer}, outliers or noisy samples will get smaller weights at $\bP$-step. For $j$-th sample, let $\bP_{\cdot j} = [\bP_{1j}, \dots, \bP_{mj}]^T$ denote its weight under each view, we can have the following observations: 1) if $j$-th sample obtains good local structures and can be reconstructed very well under each view, $\bP_{\cdot j}$ will get large values for all its entries and the self-representation under each view (i.e., $\{\bZ_{\cdot j}^v\}_{v=1}^{m}$) will all have large influence when calculating the consensus representation $\bZ_{\cdot j}^*$.  2) if $j$-th sample behaves abnormally (e.g., as outlier) only on some views, the corresponding weights in $\bP_{\cdot j}$ for these views will be very small and $\bZ_{\cdot j}^*$ is thus mainly determined by the self-representations learned from other views with high confidence levels. 
3) if $j$-th sample behaves as outliers on all the views, $\bP_{\cdot j}$ will get very small values for all its entries, and the learned $\bZ_{\cdot j}^*$ will be close to zero due to the sparse constraint in eq (\ref{problem_z_*}). 
Therefore, we can get a robust consensus representation $\bZ^*$ through fusing the noiseless structures among views and samples. 

\input{paper/algorithm}

\subsection{Convergence and Computational Complexity}
We first analyze the convergence property of Algorithm 1. During each iteration, an convex minimization problem is solved for each $\bP$-step, $\bZ^v$-step and $\bZ^*$-step, and the global optimum solution is obtained for each sub-step. Thus the overall objective is non-increasing with the iterations and Algorithm 1 is guaranteed to convergence.

For a given multi-view dataset and a fixed parameter $\lambda$ in Algorithm 1, the matrix inverse operator and operation $(\bX^v)^T\bX^v$ in eq (\ref{update_z_v}) only need to be calculated once during the optimization, and their time cost is $O(mn^3 + dn^2)$. 
In each iteration, $\bP$ is computed with the cost of $O(dn^2 + mn^2)$. For $\bZ^v$-step, there is an extra matrix multiplication operator in eq (\ref{update_z_v}), which has a time cost of $O(mn^3)$. The cost for computing $\bZ^*$ is $O(mn^2)$. Thus the total time cost of Algorithm 1 is $O(mn^3 + dn^2 + T(dn^2 + 2mn^2 + mn^3))$, where T is the total iteration number. In our experiments, the objective value decreases very fast and the convergence can be reached within a small number of iterations.

%% file: paper/algorithm.tex
\begin{algorithm}
	\caption{: Algorithm for RMSC}
	\label{alg_rmsc_ssc}
	\begin{algorithmic}[1]
		\REQUIRE Multi-view data \(\bX^{v} \in R^{d_{v}\times n}\) (\(v = 1,\dots, m \)), parameters \(\lambda\), $\beta$, number of clusters $c$.
		
		\STATE Set $t=0$, initialize sample weights $\bP(t) \in R^{m \times n}$ as all one matrix, initialize each $\bZ^v(t) \in R^{n \times n}$,  \(v = 1,\dots, m \).
		
		\REPEAT

		\STATE Update $\bZ^*(t+1)$ via eq (\ref{update_z_*}).
	
		\STATE Update $\bP(t+1)$ via eq (\ref{update_p}).
		
		\STATE Update each $\bZ^v(t+1)$ via eq (\ref{update_z_v}), \(v = 1,\dots, m \).
		
		\UNTIL {convergence}.
		
		\STATE Construct data similarity matrix $\bS = \frac{(|\bZ^*| + |(\bZ^*)^T|)}{2}$.

		\STATE Obtain clustering results by performing spectral clustering on similarity matrix $\bS$.
	\end{algorithmic}
\end{algorithm}

%% file: paper/experiments.tex
\section{Experiments}
In this section, we evaluate the proposed model in comparison with several state-of-the-art clustering methods on four real-world databases. 
Experimental results demonstrate the correctness and effectiveness of the proposed model.

\input{paper/table_acc_nmi}

\subsection{Databases} 
Four widely used real-world benchmarks are considered in the experiments. 
Their statistical information is summarized in Table \ref{statistical_information_database}.
\input{paper/table_databases}

\textbf{UCI Handwritten Digit dataset \footnote{https://archive.ics.uci.edu/ml/datasets/Multiple+Features}:} 
This dataset is taken from UCI repository. It consists of 2000 handwritten digits classified into ten categories (0-9), and each category has 200 instances. Samples are represented in six kinds of features: pixel averages in 2 x 3 windows (PIX), Fourier coefficients of the character shapes (FOU), profile correlations (FAC), Zernike moments (ZER), Karhunen-Love coefficients (KAR), and morphological features (MOR).

\textbf{Reuter Multilingual dataset \footnote{http://multilingreuters.iit.nrc.ca}:}
It contains feature characteristics of documents written in five different languages (English, French, German, Spanish and Italian), and documents in different languages share the same 6 categories. We use documents originally in English as the first view and their French, German, Spanish and Italian translations as other four views. We randomly sample 1200 documents in a balanced manner, with 200 documents in each category.

\textbf{Animal \footnote{http://attributes.kyb.tuebingen.mpg.de/.}:}
It consists of 50 kinds of animals, with 30475 images in total. The used six per-extracted features are Color Histogram, Local Self-Similarity, PyramidHOG (PHOG), SIFT, colorSIFT and SURF. 
Similar to \cite{yin2015multi}, we select the first ten categories and randomly sample 50 instances in each one as a subset for evaluation.

\textbf{3-Sources \footnote{http://mlg.ucd.ie/datasets/3sources.html.}:}
This dataset is collected from three well-known on-line news sources: BBC, Reuters and The Guardian. There are 416 distinct news stories which  are manually divided into six classes. Among them, 169 stores are reported in all three sources and are used in our experiments as in \cite{liu2013multi}. 

\subsection{Baseline Algorithms and Experimental Setting}
\label{baseline}
To better demonstrate the performance of the proposed model, we compare it with several state-of-the-art methods.

\textbf{SC-BSV:} Perform standard spectral clustering \cite{shi2000normalized} on each single-view, and the best result is presented.
\textbf{SSC-BSV:} Run SSC \cite{elhamifar2013sparse} on each view independently, and the best result is reported.
\textbf{SSC-AVG:} Run SSC \cite{elhamifar2013sparse} on each view independently to get each view's subspace representation, then perform spectral clustering on the averaged representation.
\textbf{MSC:} A weighted multi-view spectral clustering model \cite{xia2010multiview}. 
\textbf{Co-Pairwise:} A co-regularization scheme that regularizes the Laplacian embedding to have high pairwise similarity \cite{kumar2011co}. 
\textbf{Co-Centroid:} Another co-regularization scheme \cite{kumar2011co} that regularizes view-specific Laplacian embedding to be similar to a common consensus. 
\textbf{Co-Training:} A \emph{co-training} approach that alternately modifies one view's graph structure with other views information \cite{kumar2011co_training}.
\textbf{DiMSC:} A diversity induced multi-view subspace clustering method which aims to reduce the redundancy between multi-view representations \cite{cao2015diversity}. Focusing on the influence of different strategies to combine multi-view representation, we use sparse constraint on each view's self-representation instead of the original smooth regularized term for better comparison.
%
%
\textbf{RMSC:} The proposed robust localized multi-view subspace clustering model (\ref{proposed_model}).
To better demonstrate the correctness and effectiveness of the proposed RMSC model, we further design a variation of model (\ref{proposed_model}) that considers a unified weight for a view and it is named as \textbf{RMSC-WV}. RMSC-WV only distinguishes the confidence levels of views and samples in the same view will get same weights (details of its implementation are given in the appendix). 

All samples are normalized to have unit L2 norm. Parameter $\gamma$ in eq (\ref{l1_l2_regularizer}) is set to be $10^{-5}$ for all the dataset.
As k-means is applied in all the methods, we run it 20 times with random initialization and report both mean values and standard derivations. 
The Gaussian kernel and $k$-nearest neighbor graphs \cite{von2007tutorial} is used for methods that need to construct the Laplacian matrix of each view, $k$ is empirically set to be 5. 
Two commonly used metrics, i.e., clustering accuracy and normalized mutual information (NMI) \cite{chen2011parallel}, are used as evaluation measures in this paper.

\input{paper/fig_result_vary_views_digit}
\input{paper/fig_convergence_objective}
\subsection{Results and Parameter Analysis}
Table \ref{result_acc} shows the numerical results of different methods on all the four databases. It is obvious that the proposed RMSC model outperforms all the compared algorithms and can improve the performance of multi-view clustering.
According to the clustering results of SSC-AVG and SSC-BSV, we can tell that na\"ive treat each view equally and average over each view's representation cannot always boost the performance of single view clustering (e.g. SSC-AVG is worse than SSC-BSV on Digit). This is because that views with low confidence levels can have large negative influence on performance in the average operation. 
By taking into consideration of each view's confidence level and pursuing a weighted average among views, RMSC-WV can obtain consistent improvements on all the datasets over SSC-BSV model. Moreover, the proposed RMSC model can further boost the performance of RMSC-WV by simultaneously considering the confidence levels of both samples and views. This corroborates our analysis that na\"ive treating each view equally or assigning a unified weight to a view can both lead to suboptimal solutions and proves the effectiveness of the proposed RMSC model. 

To better analyze the properties of the proposed model, we further report its clustering performance on Digit and Reuter with respect to different number of used views in Figure \ref{fig_vary_views}. SSC-AVG and RMSC-WV are also implemented for comparison. 
For Digit, the views are considered in an order of PIX, FOU, FAC, ZER, KAR, MOR. For example, \{{PIX, FOU}\} and \{{PIX, FOU, FAC}\} are used when $m=2$ and $m=3$, respectively. Similarly, the views are considered in an order of English, French, German, Spanish and Italian for Reuter. 
Seen from Figure \ref{fig_vary_views}, RMSC obtains consistently better performance than SSC-AVG and RMSC-WV with respect to different number of used views on both datasets. More specifically, on Digit, as the number of used views increases, the performance of RMSC is more robust than that of SSC-AVG and RMSC-WV. 
For example, both the performance of RMSC-WV and RMSC increases when incorporating the third view (FAC). 
However, when taking into consideration of the fourth view (ZER), the performance of SSC-AVG and RMSC-WV decreases a lot while that of RMSC is more stable.
On Reuter, the performance of all the three algorithms increases as more views are included (except for the fifth view), and the RMSC always achieves the best performance. 
Thus RMSC is able to obtain robust consensus representation by considering the confidence levels of both views and samples and can improve the performance of multi-view clustering.

Figure \ref{convergence} shows the convergence performance of Algorithm 1 on Digit and Reuter. 
We can see that the overall objective value decreases very fast and the convergence can be reached within 5 iterations. 
To investigate the influence of parameter $(\lambda,\beta)$ in RMSC, we further report its clustering results with respect to $(\lambda,\beta)$ on Digit. The results are shown in Figure \ref{vary_lambda_beta}. We can observe that RMSC is not very sensitive to parameters and there exists a large parameter space of $(\lambda,\beta)$ so that it can achieve promising results.

\input{paper/fig_lambda_beta}

%% file: paper/table_acc_nmi.tex
\begin{table*}
	\small
	\centering
	\caption{Clustering performance on four benchmark databases. The best results are highlighted in bold.}
	\label{result_acc}
	\setlength\tabcolsep{3pt}
	\begin{tabular}{|c|c|c|c|c||c|c|c|c|}

		\hline
		\multirow{2}{*}{Methods} & \multicolumn{4}{c||}{Accuracy($\%$)} &  \multicolumn{4}{c|}{Normalized Mutual Information($\%$)} \\
		\cline{2-9}
		& Digit & Reuter & Animal & 3-Sources & Digit & Reuter & Animal & 3-Sources\\
		
		\hline
		SC-BSV  
		& 94.52 $\pm$ 3.48 & 33.88 $\pm$ 0.43 & 25.85 $\pm$ 1.06 & 61.89 $\pm$ 0.68 
		& 91.24 $\pm$ 0.87 & 18.60 $\pm$ 0.58 & 15.90 $\pm$ 0.61 & 61.04 $\pm$ 1.02\\
				
		\hline
		MSC
		& 94.68 $\pm$ 2.80 & 30.49 $\pm$ 3.34 & 28.39 $\pm$ 1.04 & 65.59 $\pm$ 3.62
		& 90.34 $\pm$ 1.15 & 21.87 $\pm$ 1.18 & 19.06 $\pm$ 0.84 & 68.23 $\pm$ 3.69\\
		
		\hline
		Co-Pairwise  
		& 96.16 $\pm$ 0.02 & 31.14 $\pm$ 2.75 & 25.96 $\pm$ 0.67 & 61.48 $\pm$ 0.54 
		& 91.89 $\pm$ 0.03 & 21.69 $\pm$ 0.96 & 16.08 $\pm$ 0.50 & 60.14 $\pm$ 1.54\\
		
		\hline
		Co-Centroid  
		& 96.65 $\pm$ 0.00 & 29.42 $\pm$ 2.83 & 29.52 $\pm$ 1.69 & 65.44 $\pm$ 2.60 
		& 93.71 $\pm$ 0.00 & 21.44 $\pm$ 0.85 & 19.28 $\pm$ 0.89 & 64.99 $\pm$ 1.98\\
				
		\hline
		Co-Training 
		& 85.39 $\pm$ 0.12 & 32.91 $\pm$ 1.91 & 29.63 $\pm$ 1.10 & 61.95 $\pm$ 5.30 
		& 85.44 $\pm$ 0.11 & 23.13 $\pm$ 0.75 & 18.92 $\pm$ 0.61 & 64.30 $\pm$ 2.81\\
				
		\hline
		\hline
		SSC-BSV  
		& 93.02 $\pm$ 3.42 & 49.36 $\pm$ 1.15 & 27.36 $\pm$ 0.87 & 67.16 $\pm$ 1.53    
		& 88.05 $\pm$ 1.02 & 33.27 $\pm$ 0.55 & 15.49 $\pm$ 0.62 & 57.84 $\pm$ 1.72\\
		
		\hline
		SSC-AVG
		& 85.87 $\pm$ 4.35 & 50.18 $\pm$ 1.69 & 30.10 $\pm$ 0.91 & 69.82 $\pm$ 1.89  
		& 87.89 $\pm$ 0.93 & 31.60 $\pm$ 0.70 & 17.52 $\pm$ 0.99 & 67.95 $\pm$ 1.13\\
					
		\hline
		DiMSC  
		& 90.79 $\pm$ 3.71 & 53.01 $\pm$ 1.24 & 31.83 $\pm$ 0.68 & 65.47 $\pm$ 0.98 
		& 84.46 $\pm$ 1.33 & 39.20 $\pm$ 0.21 & 19.29 $\pm$ 0.59 & 60.23 $\pm$ 2.30\\

		
		\hline
		RMSC-WV  
		& 95.35 $\pm$ 0.00 & 55.05 $\pm$ 1.06 & 31.87 $\pm$ 0.66 & 74.56 $\pm$ 1.80 
		& 90.65 $\pm$ 0.00 & 37.70 $\pm$ 0.43 & 19.70 $\pm$ 0.61 & 67.17 $\pm$ 1.95\\
		
		\hline
		RMSC  
		& \textbf{97.91 $\pm$ 0.03} & \textbf{57.50 $\pm$ 0.39} & \textbf{33.14 $\pm$ 1.01} & \textbf{78.37 $\pm$ 0.52} 
		& \textbf{94.98 $\pm$ 0.08} & \textbf{40.80 $\pm$ 0.20} & \textbf{20.02 $\pm$ 0.64} & \textbf{70.53 $\pm$ 0.71}\\
		
		
		\hline
	\end{tabular}
\end{table*}

%% file: paper/table_databases.tex
\begin{table} [tb]
	\caption{Statistical information of databases.}
	\centering
	\label{statistical_information_database}
	\begin{tabular}{|c|c|c|c|}
		\hline
		\textbf{Dataset} & \textbf{\#.Instance} & \textbf{\#.Views} & \textbf{\#. Clusters} \\
		
		\hline
		Digit & 2000 & 6 & 10 \\
		
		\hline
		Reuter & 1200 & 5 & 6 \\
				
		\hline
		Animal & 500 & 6 & 10 \\
		
		\hline
		3-Sources & 169 & 3 & 6 \\
		
		\hline
	\end{tabular}
\end{table}

%% file: paper/fig_result_vary_views_digit.tex
\begin{figure*}
	\centering
	\subfigure[Digit]{
		\includegraphics[height=1.6in]{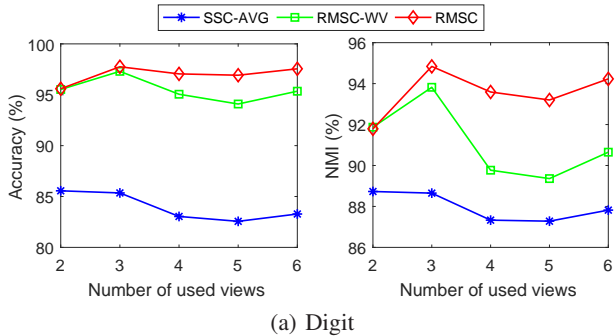}
	} \hfill
	\subfigure[Reuter]{
		\includegraphics[height=1.6in]{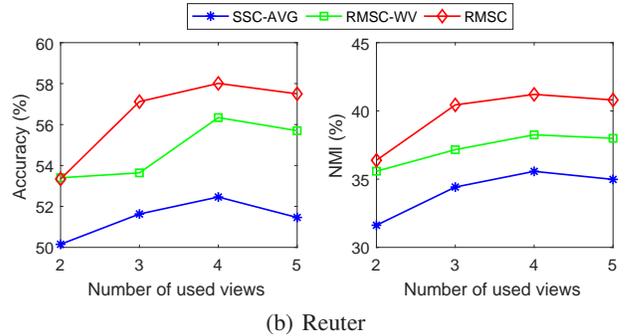}
	}
	\caption{Clustering performance on Digit and Reuter w.r.t number of used views.}
	\label{fig_vary_views}
\end{figure*}

%% file: paper/fig_convergence_objective.tex
\begin{figure}[t]
	\centering
	\subfigure[Digit]{
		\includegraphics[height=1.25in]{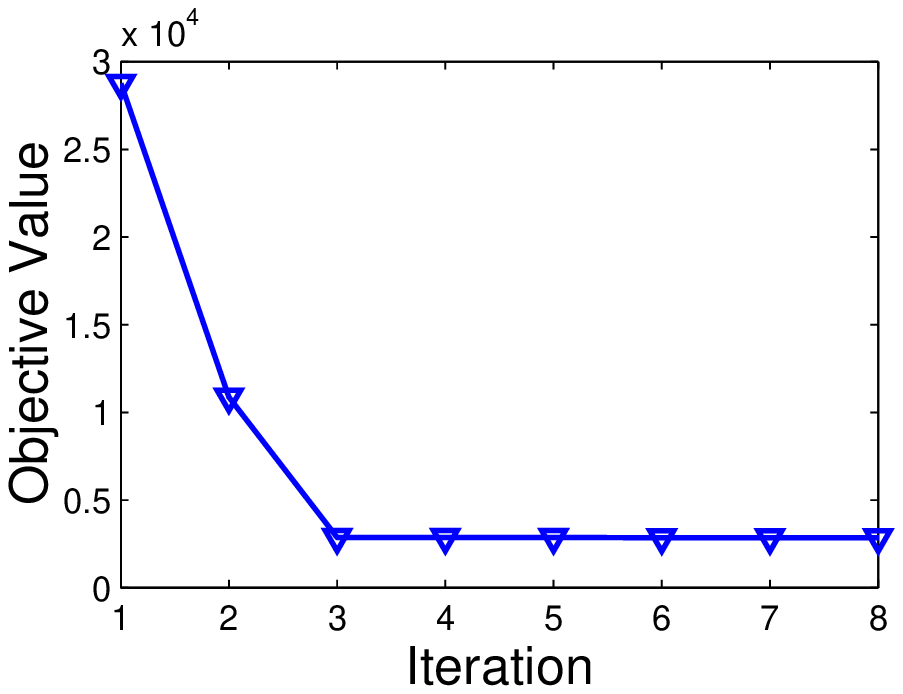}
	}
	\subfigure[Reuter]{
		\includegraphics[height=1.25in]{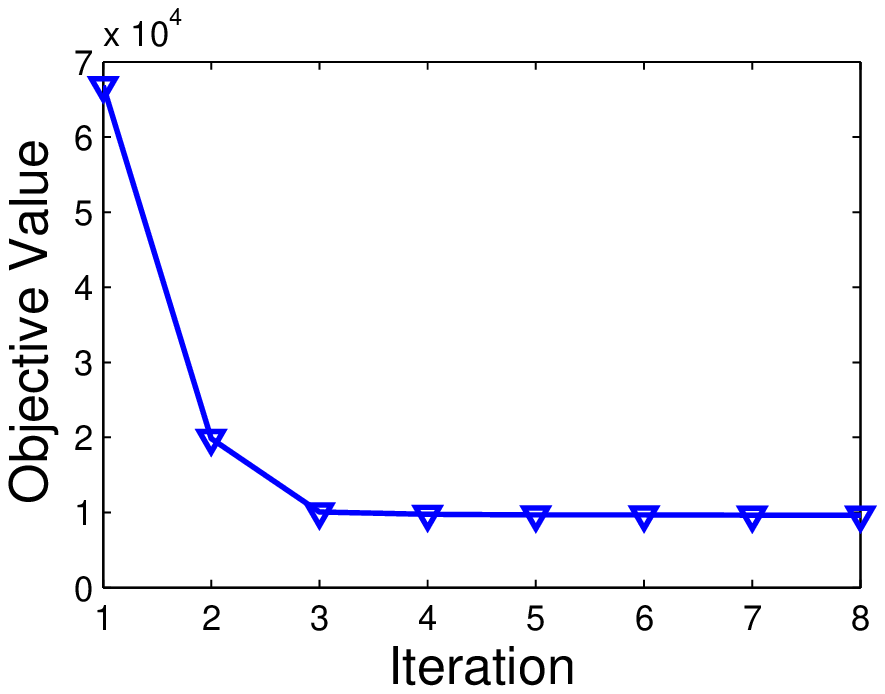}
	}
	\caption{Convergence performance on Digit and Reuter.}
	\label{convergence}
\end{figure}

%% file: paper/fig_lambda_beta.tex
\begin{figure}
	\centering
	\subfigure{
		\includegraphics[height=1.4in]{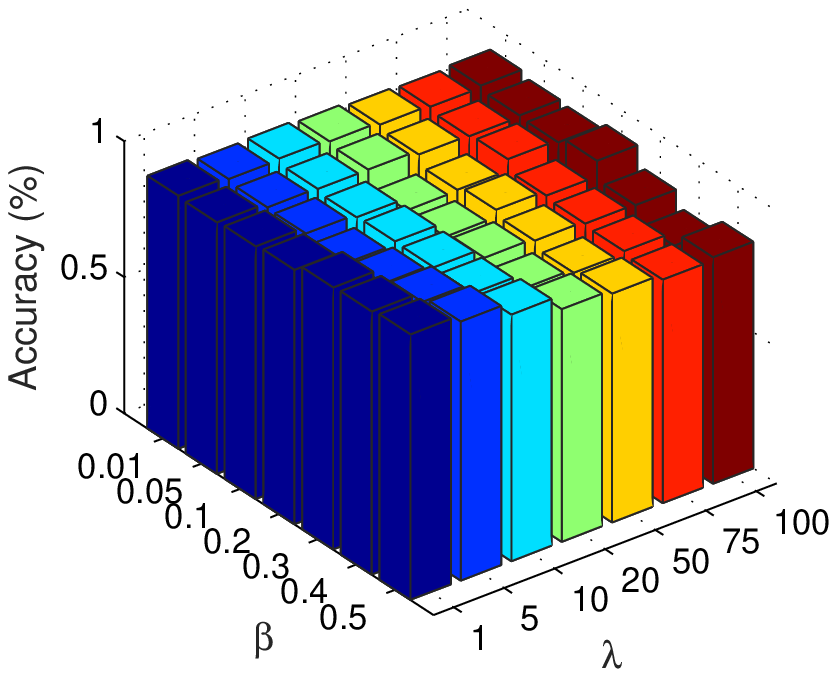}
	}
	\subfigure{
		\includegraphics[height=1.4in]{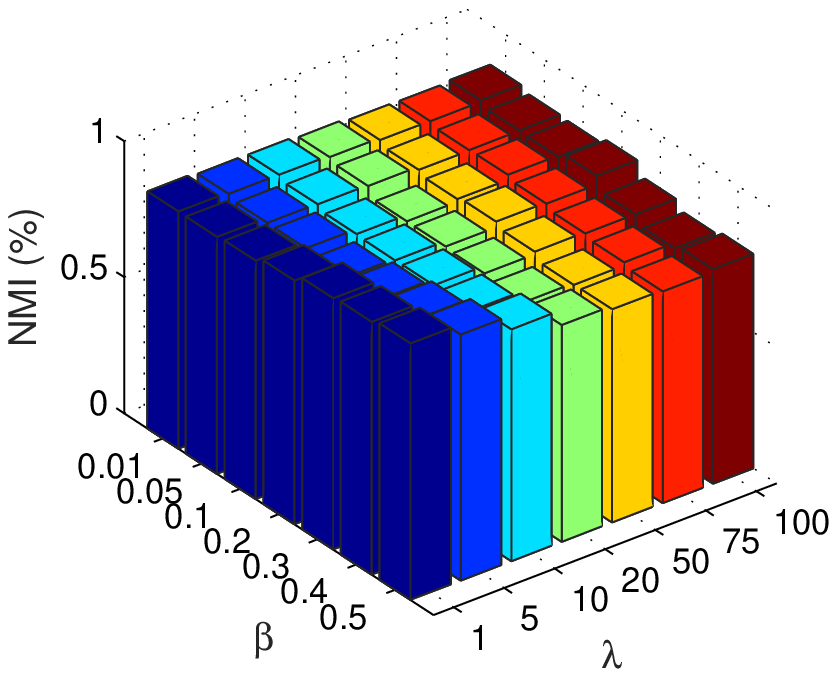}
	}
	\caption{Clustering performance on Digit w.r.t. $\lambda$ and $\beta$}
	\label{vary_lambda_beta}
\end{figure}

%% file: paper/conclusion.tex
\section{Conclusion}
In this paper, we have proposed a novel robust localized multi-view subspace clustering (RMSC) model that considers the confidence levels of both samples and views. RMSC aims to learn a consensus self-representation for multi-view data, and the proposed weighting strategy can reflect the samples confidence levels to some extent. We further develop an iterative optimization method for RMSC, and it converges quickly in few iterations. Comprehensive experimental results on four benchmark datasets show that RMSC can obtain a robust consensus representation and outperforms the state-of-the-art multi-view clustering algorithms. 


%% file: paper/appendix.tex
\section{Appendix}

RMSC-WV is a variation of model (\ref{proposed_model}) and it considers a unified weight for each view (i.e., samples in the same view will be assigned same weight). The formulation of {RMSC-WV} defined in Section \ref{baseline} is 
\begin{equation}
\small
\begin{aligned}
& \min_{\{\bZ^v\}_{v=1}^m,\bZ^*,\bp}  \ \sum_{v=1}^{m} \{\bp_{v} (||\bX^v - \bX^v \bZ^v||_F^2 + \lambda ||\bZ^v - \bZ^*||_F^2)  \\ 
& \quad \qquad \qquad \qquad \qquad + \psi(\bp_{v})\}  + \beta||\bZ^*||_1 \\
& \ \ s.t. \quad \bp \in R^{m}, \ \bp_v \geq 0, \ diag(\bZ^v) = 0, v=1,\dots,m. \\
\end{aligned}
\label{rmsc_mv}
\end{equation}
where $\bp_v$ denotes the weight of $v$-th view and $\psi(\cdot)$ is a regularizer on $\bp_v$. $\psi(\cdot)$ is defined in eq (\ref{l1_l2_regularizer}). Similar to model (\ref{proposed_model}), block coordinate descent strategy is used for its optimization.